\documentclass[sigconf]{acmart}
\usepackage{amsmath}

\copyrightyear{2020} 
\acmYear{2020} 
\setcopyright{rightsretained} 
\acmConference[SIGIR '20]{Proceedings of the 43rd International ACM SIGIR Conference on Research and Development in Information Retrieval}{July 25--30, 2020}{Virtual Event, China}
\acmBooktitle{Proceedings of the 43rd International ACM SIGIR Conference on Research and Development in Information Retrieval (SIGIR '20), July 25--30, 2020, Virtual Event, China}\acmDOI{10.1145/3397271.3401433}
\acmISBN{978-1-4503-8016-4/20/07}

\DeclareMathAlphabet\mathbfcal{OMS}{cmsy}{b}{n}

\begin{document}
\fancyhead{}

\title{Efficient Image Gallery Representations at Scale Through Multi-Task Learning}

\author{Benjamin Gutelman}
\email{benjamin.gutelman@booking.com}
\affiliation{%
  \institution{Booking.com}
  \city{Tel Aviv}
  \country{Israel}
}
\author{Pavel Levin}
\email{pavel.levin@booking.com}
 \affiliation{%
  \institution{Booking.com}
  \city{Tel Aviv}
  \country{Israel}
}

\renewcommand{\shortauthors}{Gutelman and Levin}
\newcommand{\TODO}[1]{\textbf{ X}\footnote{TODO: #1}}
\begin{abstract}
  Image galleries provide a rich source of diverse information about a product which can be leveraged across many recommendation and retrieval applications. We study the problem of building a universal image gallery encoder through multi-task learning (MTL) approach and demonstrate that it is indeed a practical way to achieve generalizability of learned representations to new downstream tasks. Additionally, we analyze the relative predictive performance of MTL-trained solutions against optimal and substantially more expensive solutions, and find signals that MTL can be a useful mechanism to address sparsity in low-resource binary tasks.
\end{abstract}



\begin{teaserfigure}
  \hspace{-5pt}
  \includegraphics[width=\textwidth]{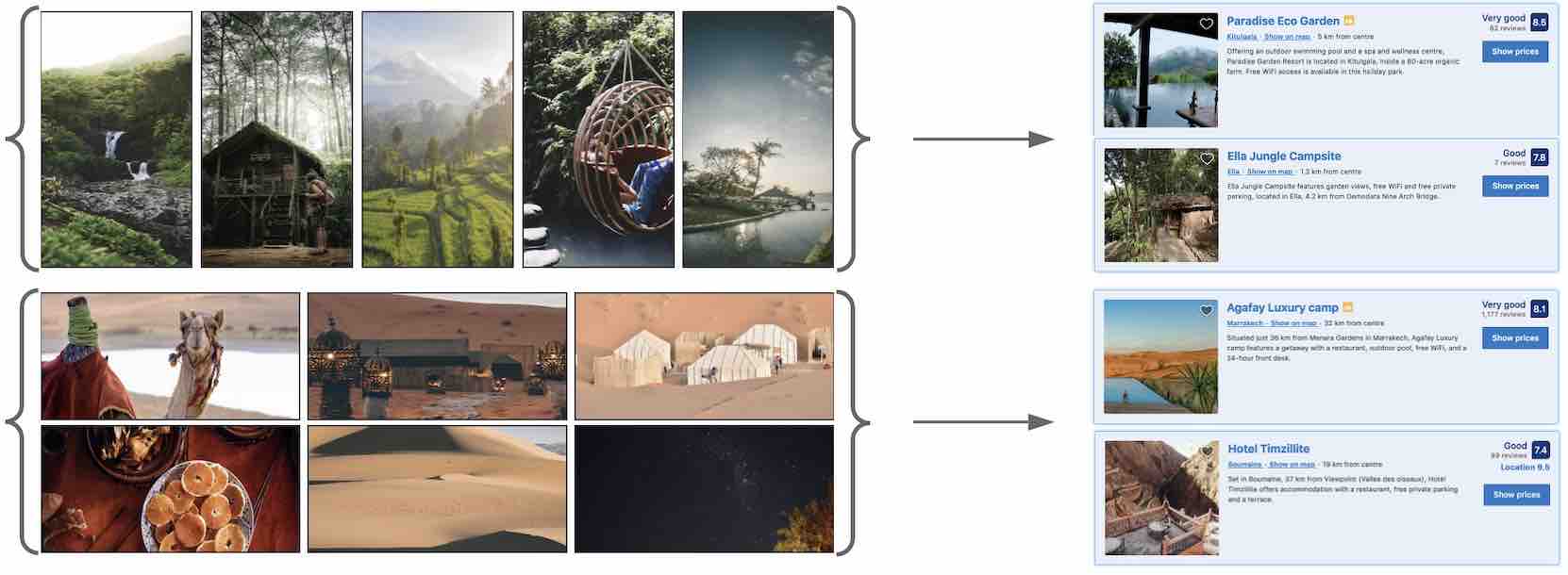}
  \caption{Example of a retrieval task based on gallery representations:  accommodations are globally ranked according to the proximity of their photo gallery embeddings to the gallery of ``seed" images. Multi-task training ensures that the embeddings capture business-relevant product attributes and not just image features.}
  \label{fig:retrieval}
\end{teaserfigure}

\maketitle

\section{Introduction}
Today's web experience is becoming increasingly more visual. A product description in a typical e-commerce marketplace is usually accompanied by a related image gallery. Whether those images are uploaded by the sellers themselves, or sourced by the e-commerce platform, they tend to considerably enrich user experience and provide customers with a valuable informational resource at various stages of their decision-making process \cite{di2014picture, ma2019understanding}.

\begin{figure}
  \centering
  \includegraphics[width=.49\linewidth]{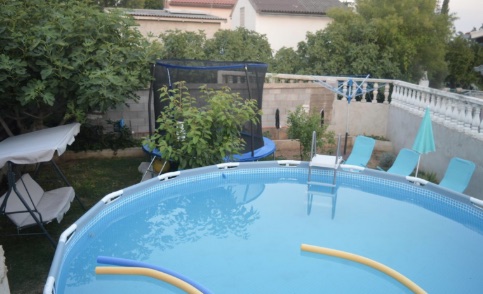}
  \includegraphics[width=.49\linewidth]{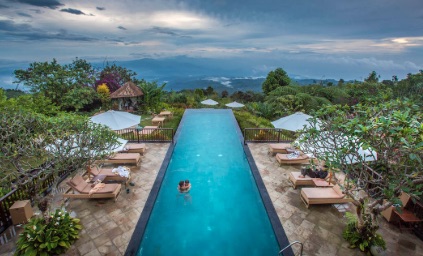}
  \caption{Swimming pools from two different accommodations. While many qualifying details and nuances can be easily inferred from the photos by website visitors, those details are challenging to represent directly in a structured and usable format.}\label{fig:good_bad_pool}
\end{figure}

A person looking through such a gallery can usually infer a great deal about the product, often even more than they can from textual descriptions or formatted specifications. For instance, in travel domain accommodation features like the quality of a swimming pool (Figure \ref{fig:good_bad_pool}), or of the room view can be subjective and difficult to represent directly in a generalizable way. The problem of finding a meaningful representation for the entire photo gallery clearly becomes even more complex.

This work describes a deep learning-based solution to the problem of finding meaningful representations of image galleries in a large-scale e-commerce setting. The solution is designed to satisfy three important constraints: (i) invariance of the representations under gallery re-orderings, (ii) their universality and flexibility to new downstream tasks, (iii) feasibility to deploy the solution in a large-scale e-commerce setting. The first requirement is essentially a consistency requirement which says that the information we obtain from a set of images is independent of the order in which we process these images. We enforce this by explicitly designing the gallery encoder as a symmetric function. The second constraint is addressed by training the gallery encoder on multiple independent tasks through a multi-task learning approach. The third requirement reflects the fact that this research problem emerged from real-world information retrieval and recommendation needs, and it constrains us to favour faster and more scalable architectures.

\section{Background}

\subsection{Image representations}
Recent advances in deep learning, particularly convolutional neural networks (CNN) enabled impressive performance gains in a number of challenging visual recognition tasks \cite{krizhevsky2012imagenet,  szegedy2015going, he2016deep}. These successes led to a widespread industrial adoption. While the most direct application of a CNN classification model is to use its output to categorize or label photos, we can also use its intermediate layer activations to represent images in a semantically meaningful way. These representations are commonly referred to as \textit{image embeddings}. Intuitively, a deep neural classifier loads raw pixel intensities and then extracts increasingly abstract representations through its layers, which eventually lead to a classification attempt in the space of high-level concepts. Therefore, it is forced to learn meaningful features without any direct supervision on what those features should be, and it is reasonable to expect them to be useful for related tasks.

It is important to note, that these representations are always limited to the domain in which the original model was trained. For example, representations obtained from a model trained on general domain ImageNet \cite{imagenet_cvpr09} data alone, would not be perfectly suitable to travel-related queries (e.g. price, location, ``family-friendliness", distance from landmarks, etc). To address such issues we often need to adapt, or fine-tune, the original model to be more suitable to the target domain. A common way to address such domain adaptation problem is by freezing part of the model parameters (usually lower-level layers), and allowing the rest of the network to be trained on in-domain data. That way we do not need to re-learn many of the lower-level features (which may stand for edges, gradients, basic shapes, etc.), but instead focus on learning relevant higher-level abstractions \cite{oquab2014learning}.

\subsection{Representing sets}
Since image collections are just sets of individual images, any function processing them is by definition a set function. In practical terms, we want such encoding function to be permutation invariant, i.e.\ different re-orderings of the same set of input images should produce the same output.

Recently there has been a surge of work related to machine learning on set-valued inputs. Despite a significant portion of that work being motivated by specific applications, such as point cloud data, many of the theoretical results and proposed architectures are of general scope. The theory behind learning set representations is closely related to statistical theory of exchangeability \cite{korshunova2018bruno, zaheer2017deep}. Attention-based architectures have also been extended to accommodate set learning setting \cite{vinyals2015order, lee2019set}. It is worth noting that although attention-based architectures have already proven very powerful in several domains (particularly NLP \cite{Vaswani2017AttentionIA, devlin2018bert}), we leave them outside the scope of this work mainly due to the practical requirement of fast model retraining and online retrieval times. Additionally, Zaheer et al.\ \cite{zaheer2017deep} provide some theoretical reassurance that conceptually simpler approaches can in principle provide sufficient expressive power.

\subsection{Multi-task learning}
Multi-task learning (MTL; \cite{caruana1997multitask, zhang2018overview}) refers to an umbrella of machine learning strategies which incorporate training signals from multiple feedback sources. MTL is attractive in many practical settings because of its inductive bias to generalize to new tasks. This in turn boosts computational---and often statistical---efficiency of machine learning solutions. MTL methods in deep learning setting vary according to how they share learned knowledge between tasks, and can be roughly split into two broad categories: soft and hard parameter sharing models \cite{ruder2017overview}. Soft parameter sharing approaches keep separate models for each tasks, but regularize models to be similar to one another in some sense (e.g.\ by adding an $L^2$ distance between corresponding parameters of different models to the overall loss function \cite{duong-etal-2015-low}). Alternatively, hard parameter sharing techniques directly reuse the same model parameters for different tasks. Because our goal is to build a single universal encoder, soft parameter sharing techniques are not directly applicable in our setting, and we focus on hard parameter sharing approaches instead. Within the hard parameter sharing approaches we also must limit ourselves to the architectures which naturally support a single encoder, thus some of the newer interesting techniques which leverage several encoders \cite{standley2019tasks} are also not relevant for us. This work will focus exclusively on an architecture with a single encoder with multiple decoders/heads.

In order to train an MTL system, the overall loss function needs to combine individual tasks' losses. The way it is typically achieved is by defining the overall loss as a weighted average of individual losses \cite{gong2019comparison}. Weighting individual losses is a difficult and largely unsolved problem in the industry \cite{karpathy}. One promising direction is to use adaptive weighting scheme according to each task's homoscedastic uncertainty \cite{kendall2018multi}, a strategy which we empirically evaluate in this paper.

\subsection{Related work}
There have been a number of applications involving symmetric gallery encoders in recent years. Permutation-invariant encoders using memory networks \cite{weston2014memory} have recently been successfully applied in medical imaging domain, classyfying cancer types from collections of histopathology images \cite{kalra2019learning}. Zaheer et al.\ \cite{zaheer2017deep} applied their sum-decomposition method to several applications involving collections of images: vision-based arithmetics and outlier detection.

Multi-task learning has also been used extensively in computer vision applications. For example, R-CNN models \cite{ren2015faster} use multi-task learning to jointly predict object classes and locations of the bounding bosses, thus mixing both classification and regression losses. Kendall et al.\ \cite{kendall2018multi} introduced  uncertainty-based loss weighting strategy which was used to jointly learn semantic and instance segmentation as well as per-pixel depth, thus also successfully combining different types of losses. For more examples of MTL applications in computer vision domain see \cite{teichmann2018multinet, eigen2015predicting, kokkinos2017ubernet, misra2016cross, zhang2018overview}.

To the best of our knowledge there is no prior work which applies multi-task learning to obtain generalizable representations of sets of images.

\section{Methodology}
\begin{figure}[h]
  \centering
  \includegraphics[width=\linewidth]{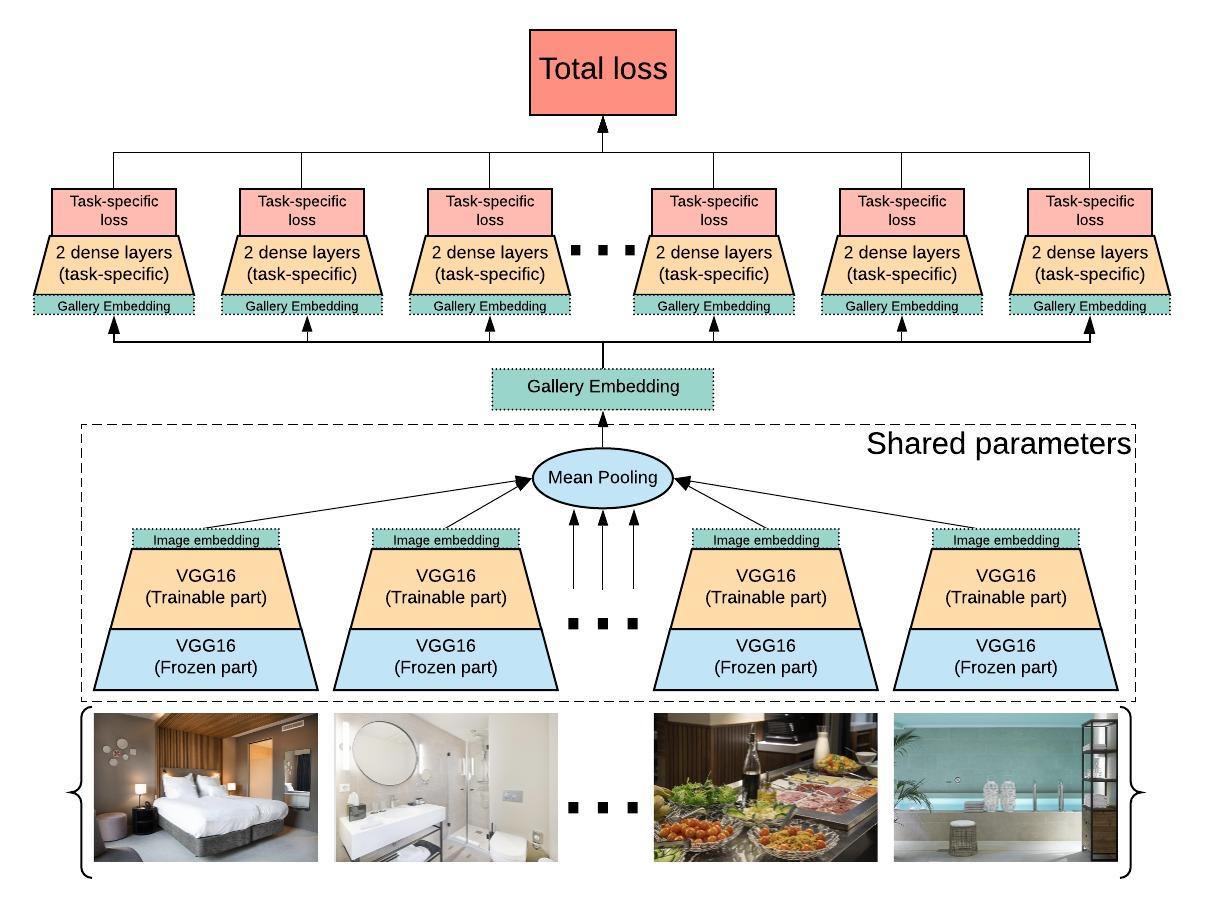}
  \caption{MTL approach to learning image gallery representation. Shared part of the model includes image encoder and the embeddings aggregation mechanism (VGG16 encoder and simple mean pooling in our case). The resulting gallery representation is then fed into task-specific 2-layer networks. Individual losses are combined into the total loss through a MTL approach. All trainable parameters (shown in yellow) are learned jointly.}\label{fig:architecture}
\end{figure}

This section describes our data, model specifications, and our training procedure. On the high level, in order to encode an image gallery we first turn each image into a vector using an image encoder (shared across all input images). Then the set of image representation vectors is passed into a permutation-invariant function computing the overall gallery representation. Gallery representations are then fed into task-specific ``heads" defined by shallow models with associated loss functions. Finally, we train the entire structure jointly in a multi-task setting (see Figure \ref{fig:architecture}).

\subsection{Dataset}
In our work we use a proprietary dataset of 520,000 accommodations sampled from Booking.com listings, randomly split into train and development sets (9:1 split). Each accommodation comes with a photo gallery and 31 additional features which we use as training signals in our multi-task learning. Table \ref{tab:features} summarizes accommodation features which we use as outputs for model training.

The lengths of accommodations' photo galleries range between 5 and 44, depending on the number of available images. The additional features, which we use as training tasks, are a mix of 17 binary classifications, 7 non-binary classifications (number of classes range between 6 and 1981) and 7 regression tasks with scales spanning distances in km, prices in euros, and values of review scores.

\begin{table}[h]
    \centering
    \begin{tabular}{r|r|l}
        \textbf{Task}& \textbf{Description}   & \textbf{Cardinality}\\
        \toprule
        C1 & City (for top destinations) & 1981\\
        C2 & World region & 9\\
        C3 & Country  & 107\\
        C4 & Type (e.g.\ Hotel, B\&B, etc.) & 28\\
        C5 & Size (bucketized \# of rooms) & 9\\
        C6 & Hotel segment & 6\\
        C7 & Vacation rental type & 6\\[3pt]
        \hline
        B1 & Is part of a hotel chain & Binary\\
        B2 & Is vacation rental  & Binary\\
        B3 & Offers sea view & Binary\\
        B4 & Offers city view & Binary\\
        B5 & Offers mountain view & Binary\\
        B6 & Is in a block of flats & Binary\\
        B7 & Has a balcony & Binary\\
        B8 & Has a kitchen & Binary\\
        B9 & Has a bar & Binary\\
        B10 & Has a garden & Binary\\
        B11 & Access to the beach & Binary\\
        B12 & Has a picnic area & Binary\\
        B13 & Has a communal lounge & Binary\\
        B14 & Offers hiking options & Binary\\
        B15 & Has a swimming pool & Binary\\
        B16 & Has a fitness center & Binary\\
        B17 & Is in the city center & Binary\\[3pt]
        \hline
        R1 & Distance from city center  & Number (km)\\
        R2 & Review score  & Number (0-10)\\
        R3 & Average daily rate  & Number (euros)\\
        R4 & Location score  & Number (0-10)\\
        R5 & Content score  & Number (0-1)\\
        R6 & Average booking window  & Number (days)\\
        R7 & Average length of stay  & Number (days)\\
        
    \end{tabular}
    \caption{Accommodation features which we use as learning targets in our MTL set up.}
    \label{tab:features}
\end{table}

\subsection{Model architecture}\label{sec:model_arch}
The overall model consists of shared and task-specific components. The shared part is a VGG16 encoder \cite{simonyan2014very} for individual images followed by the mean pooling function applied across all the image embeddings. Given an input gallery of $N$ images (decoded to their RGB representations and resized using bilinear interpolation to $224\times 224 \times 3$ dimension), the VGG16 encoder produces $N$ vectors of size $512$ which are reduced to an averaged vector via mean pooling. Note that symmetry under gallery re-orderings can be preserved using much more sophisticated mechanisms than just average pooling (see e.g.\  \cite{kalra2019learning, zaheer2017deep, vinyals2015order}), however those encoders come at a significant computational cost in a large-scale industrial system. We leave the detailed study of the relative benefits of using more advanced permutation-invariant encoders for future research.

The non-shared part of the model is a collection of task-specific learners. Each learner is a shallow neural network which takes 512 inputs from the gallery encoder described above, passes them through a single fully-connected layer with ReLU activation, followed by an output layer with appropriate activation (sigmoid, softmax or linear, for binary, multiclass or regression tasks respectively).

The overall model contains $12.8$M trainable parameters: $7.1$M belong to VGG16 image encoder and $5.7$M to individual task learners. Note that VGG16 encoder consists of five VGG-blocks of convolutional layers, which have $14.7$M parameters in total, however we freeze the first four blocks (roughly half of the weights) and only leave the last block as trainable, which is the reason for $7.1$M trainable parameters in the image encoder component.

\subsection{Loss function}\label{sec:loss}
As we are in a multi-task setting, we need to take special care of the loss function. Given $n$ distinct tasks, the joint loss can be written as a vector:
\begin{multline*}
    \mathbfcal{L} \left( \widehat{y_1}, \ldots, \widehat{y_n},\mathbf{y_1}, \ldots, \mathbf{y_n} \right)\\
 \mathbf{\ \ \ \ \ }
 =\left [\mathcal{L}_1\left(\mathbf{\widehat{y}}_{1} \left (\mathbf{w}_{shared},\mathbf{w}_{1}\right),\mathbf{y}_{1}\right),\ldots,\mathcal{L}_n\left(\bold{\widehat{y}}_{n} \left (\mathbf{w}_{shared},\mathbf{w}_{n}\right),\mathbf{y}_{n}\right)\right ]
\end{multline*}

The overall loss vector contains $n$ components, each corresponding to a task. $\mathcal{L}_i$ is either a categorical cross-entropy or mean squared error (MSE) function depending on the nature of the task. For any given data point $\widehat{y_i}$ denotes the predicted value of task $i$ under model parameters $(\mathbf{w}_{shared}, \mathbf{w_i})$, while $\mathbf{y_i}$ is the ground truth. In order to find optimal model parameters $\mathbf{W} = \left (\mathbf{w}_{shared},\mathbf{w_1}, \ldots, \mathbf{w_n} \right )$, we reduce the loss vector to a scalar-valued loss function $\mathcal{L}^{tot}$ using one of the two weighting procedures described below.

Our first method is based on a common uniform scaling heuristic which assigns equal weight to each task's loss. While we assign equal weights of $1$ to each classification loss, we modify the heuristic to add a constant scaling factor to the regression MSE losses as a simple way to adjust for the different scales of variation. This factor is chosen to be the inverse of the sample variance of the given regression variable in the training dataset. The minimization objective under this method (which we refer to as pseudo-uniform scaling) is the following:
\begin{align*}
\mathcal{L}_{tot}&= \sum_{i\, \in\, cat,bin  }\mathcal{L}_{i} + \sum_{j\, \in\, reg}\frac{1}{\sigma_{j}^{2}}\mathcal{L}_{j}\\
 &= -\sum_{i\, \in\,  cat }\mathbf{y_{i}}\log\bold{\widehat{y}_{i}} -\sum_{j\, \in \, bin }y_{j}\log{\widehat{y}_{j}}
+\left( 1-y_{j}\right )\log\left(1-\widehat{y}_{j} \right )\\
&\mathbf{\ \ \ \ \ \ \ \ \ \ \ \ \ \ \ \ \ \ \ \ \ \ \ \ \ \ \ \ \ \ \ \ \ \ \ \ \ \ \ \ \ \ \ \ \ \ \ \ \ \ \ \ \ \ \ \ }
+\sum_{k\, \in\, reg}\sigma_{k}^{-2}\lvert\widehat{y}_{k}-y_{k} \rvert^{2}
\end{align*}
where $\sigma_k^{2}$ is the sample variance of a regression label $k$, and $bin$, $cat$, $reg$ denote the sets of binary, multiclass and regression tasks respectively.

The second approach follows Kendall et al.\ \cite{kendall2018multi} and uses homoscedastic uncertainty to assign relative weights to each output. In this method the scaling factors are dynamic, and are learned jointly with the model parameters:
\begin{align*}
    \mathcal{L}_{tot} &= \sum_{i\, \in \, cat,bin  } \frac {1}{\sigma_{i}^2}\mathcal{L}_{i} + \sum_{j\, \in\, reg}\frac{1}{2\sigma_{j}^{2}}\mathcal{L}_{j}+\sum_{k\, \in \, cat,bin,reg}\log\sigma_{k}\\
    &= -\sum_{i\, \in \, cat }\sigma_{i}^{-2}\mathbf{y_{i}}\log\mathbf{\widehat{y}_{i}} -\sum_{j\, \in \, bin }\sigma_{j}^{-2}\Big[y_{j}\log{\widehat{y}_{j}}\\
    &\mathbf{\ \ \ \ \ \ }
    +\Big( 1-y_{j}\Big)\log\left(1-\widehat{y}_{j} \right ) \Big] +\frac{1}{2}\sum_{k\, \in \, reg}\sigma_{k}^{-2}\lvert\widehat{y_{k}}-y_{k} \rvert^{2}  \\
    &\mathbf{\ \ \ \ \ \ \ \ \ \ \ \ \ \ \ \ \ \ \ \ \ \ \ \ \ \ \ \ \ \ \ \ \ \ \ \ \ \ \ \ \ \ \ \ \ \ \ \ \ \ \ \ \ \ \ \ }
    + \sum_{l\, \in \, cat,bin,reg}\log\sigma_{l}
\end{align*}
Unlike in the pseudo-uniform scaling case, each $\sigma_k^2$ is not directly calculated from data, but is instead a learned parameter.

\subsection{Model training}
The training pipeline is implemented with Tensorflow v1.12.0 \cite{tensorflow2015-whitepaper} and executed on 32 vCPU's and a single NVIDIA P100 GPU. The vCPU's are used in parallel to load and preprocess the binary image data stored in TFRecords format, while the GPU is used to execute machine learning training passes over the transformed data batches. We use Adam optimizer \cite{kingma2014method} with $\beta_1=0.9$, $\beta_2=0.999$ and learning rate $1.0\times 10^{-4}$ for model training, which is run until the total loss in our validation set stops decreasing. This took roughly four epochs ($1.6\times 10^{6}$ iterations) for both versions of the total loss functions as described in Section \ref{sec:loss}. Each data point may include up to 44 images, which limited us to the maximum batch size of 4.

\begin{figure*}
  \includegraphics[width=\textwidth]{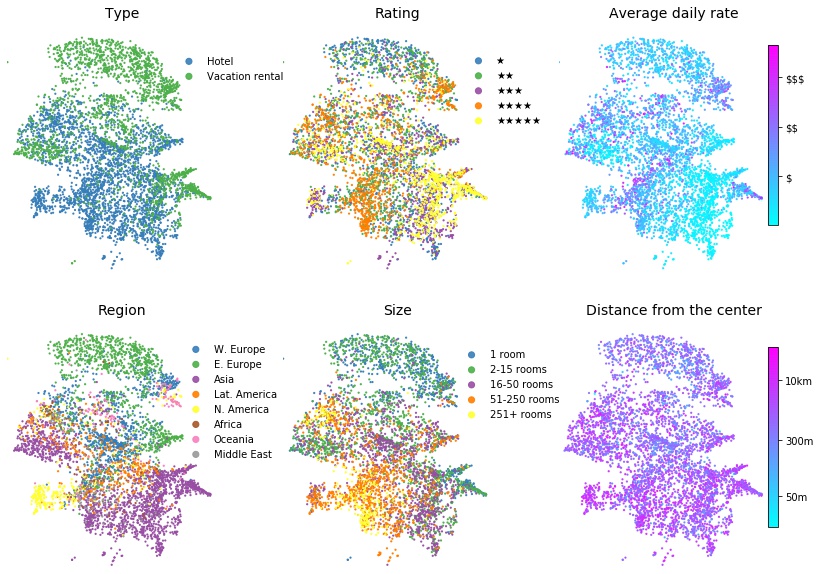}
  \caption{2D projections of property representations given by image gallery embeddings. The visualizations demonstrate that the encoder learned to extracts meaningful domain-related features from photo galleries.}
  \label{fig:embeddings}
\end{figure*}

\section{Evaluation}
Our work is motivated by many real-world recommendation and retrieval tasks which require some form of product representation. We hypothesize that image gallery alone provides a rich source of information, and that we can use MTL to learn how to compress visual gallery into useful features for downstream tasks. As such, our main evaluation criterion is how well our gallery encoding approach can extract features useful to \textit{new} downstream tasks. To evaluate this we look at the gains in computational and statistical efficiency in a ``hold out" task of predicting property star rating from image gallery (Section \ref{sec:holdout}). Section \ref{sec:comparison} analyses relative performance of the two MTL approaches. 

Figure \ref{fig:retrieval} and Figure \ref{fig:embeddings} provide limited qualitative evaluation of the embeddings. The former is a visualization of two examples of embeddings-based accommodation retrieval based on a collection of ``seed" photos. The fast similarity search was performed using bucketed random projections version of the locality-sensitive hashing \cite{indyk1998approximate}. Figure \ref{fig:embeddings} offers a 2D mapping of representations \cite{mcinnes2018umap} learned by the gallery encoder on a random sample of accommodations.

\subsection{New task: star rating prediction}\label{sec:holdout}
We judge the quality of learned gallery embeddings by how generalizable they are to new downstream tasks. To evaluate this, we consider a ``hold out" task of predicting star ratings of accommodations. The star rating feature was present in $250$k properties in our dataset, but we did not use it in any way when training the overall MTL model\footnote{Because we saw pseudo-uniform scaling outperforming Kendall et al.\ \cite{kendall2018multi} on most training task in terms of individual in-task performance (Section \ref{sec:comparison}), our evaluation in this section as well as visualizations (Figures \ref{fig:retrieval} and \ref{fig:embeddings}) are based on pseudo-uniform MTL}. Table \ref{tab:star_rating} compares the accuracy of our model against two baselines. The first baseline (``Imagenet transfer")\, substitutes the representations from our gallery encoder with averaged photo embeddings, where photo embeddings are obtained from a VGG16 model fully pretrained on ImageNet \cite{imagenet_cvpr09} data. The second baseline (``End-to-end model") is a model of the same configuration as described in Section \ref{sec:model_arch}, but trained end-to-end on one specific objective of classifying property star rating.

\begin{table}[]
\resizebox{1\linewidth}{!}{
\begin{tabular}{r|ccc}
\multicolumn{1}{l}{}  & \multicolumn{1}{c}{\begin{tabular}[c]{@{}c@{}}MTL\\ embeddings\end{tabular}} & \multicolumn{1}{c}{\begin{tabular}[c]{@{}c@{}}ImageNet\\ transfer\end{tabular}} & \multicolumn{1}{c}{\begin{tabular}[c]{@{}c@{}}End-to-end\\ model\end{tabular}} \\
\toprule
50k (20\% of data)    & \textbf{0.574}& 0.515& 0.543\\
100k (40\% of data)   & \textbf{0.597}& 0.523& 0.558\\
150k (60\% of data)   & \textbf{0.616}& 0.546& 0.575\\
200k (80\% of data)   & \textbf{0.622}& 0.532& 0.586\\
250k (a full epoch)   & \textbf{0.623}& 0.534& 0.583\\
1000k (four epochs)   & \textbf{0.631}& 0.542& 0.598\\
\midrule
Mean time per instance     & \multicolumn{2}{c}{14 ms (on CPU)}& \multicolumn{1}{c}{136 ms (GPU)}\\
Total time (4 epochs) & \multicolumn{2}{c}{4 hrs (on CPU)}& \multicolumn{1}{c}{40 hrs (GPU)}
\end{tabular}}
\caption{Accuracy results of the hold out star rating prediction task. Using our MTL-trained embeddings consistently outperforms training the model end-to-end or using averaged ImageNet-trained embeddings.}\label{tab:star_rating}
\end{table}

As we can see in Table \ref{tab:star_rating}, our solution consistently outperforms both alternative approaches, even on relatively small data sets. The somewhat surprising fact that even end-to-end solution lags behind MTL embeddings shows that MTL helps with learning relevant features which generalize beyond just the training tasks. Figure \ref{fig:good_bad_pool} can provide some intuition as to why ImageNet transfer does not perform as well on our task: if the original model's objective is only concerned with object classification, then it will ignore many of the characteristics which are relevant to describing the object's quality.

In terms of statistical efficiency, the end-to-end model only catches up to MTL embeddings-based model trained on 50k data points when it has 3 times as much data (150k). In terms of computational efficiency, the end-to-end model trained on NVIDIA P100 GPU and took 10 times more time per iteration then the MTL transfer model trained on CPU.

\subsection{Performance on individual tasks}\label{sec:comparison}
\begin{figure}
    \centering
    \includegraphics[width=\linewidth]{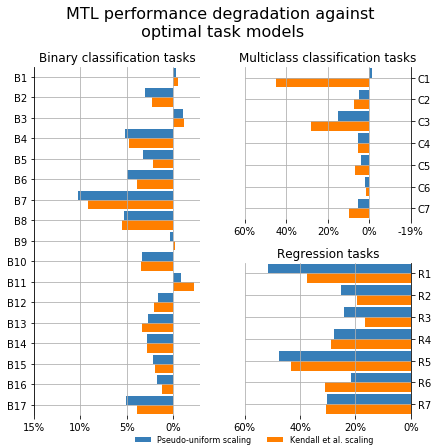}
    \caption{Relative performance of MTL against task-optimal models. We measure performance degradation in terms of error increase (accuracy error for classification and mean squared error for regression tasks). Most binary tasks are within 5\% relative difference against the optimal models. Multiclass and regression tasks generally do worse. Tasks B1, B3, B11 and C1 (for pseudo-uniform scaling only) actually do better in the MTL setup.}
    \label{fig:tasks}
\end{figure}

Even though the main purpose of training multiple tasks simultaneously is to enrich our universal gallery representations to improve their potential for generalization, it is still interesting to compare the performance of each of the individual tasks against the optimal task-specific models\footnote{Optimal task-specific models share the same architecture as described in Section \ref{sec:model_arch}, but instead trained end-to-end on their respective objectives}. Figure
\ref{fig:tasks} summarizes the results. We look at two MTL approaches (pseudo-uniform scaling, and Kendall et al.\ loss \cite{kendall2018multi}) and their predictive performance relative to optimal models individually trained end-to-end for each of the tasks.

While most tasks understandably show inferior predictive performance relative to the very strong baseline of optimal task-specific models, some tasks actually seem to benefit from MTL set up. Specifically C1, a classification task with the highest label cardinality, exhibits roughly the same performance as the optimal model under pseudo-uniform scaling set up.  Additionally, tasks B1, B3 and B11 perform better than the optimal models for both MTL total loss types. None of the regression tasks show better performance under MTL training. In general, we see pseudo-uniform scaling to do better than Kendall et al. \cite{kendall2018multi} method in 22 out of 31 tasks (71\%). This is somewhat consistent with Senter et al.\ \cite{sener2018multi} who also found uniform heuristic to outperform uncertainty-based weighting in some MTL settings.

\begin{figure}
    \centering
    \includegraphics[width=\linewidth]{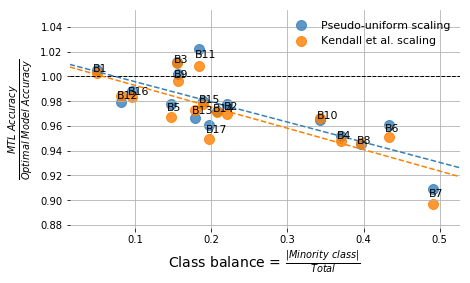}
    \caption{The relationship between class balance and MTL benefit. It seems that MTL approach degrades the predictive performance less and even provides improvement in situations of higher class imbalance. The linear relationship is statistically significant: Pearson's $r=0.78$ ($p <.001$) and $r=0.80$ ($p <.001$) in pseudo-uniform MTL and Kendall MTL respectively.}
    \label{fig:imbalance}
\end{figure}

As most of our tasks are binary classification tasks, we zoom in on how MTL compares to task-specific models as a function of class balance, which we define as the fraction of the minority class labels in the feature (e.g.\ a perfectly balanced variable will be at $0.5$, while tasks with significant class imbalance will be closer to $0$). Figure \ref{fig:imbalance} shows a clear trend for both types of MTL strategies that we tried. In terms of statistical performance, MTL seems to bring the most value to tasks with the highest class imbalance. We hypothesise that this is happening because more skewed tasks benefit more from additional learning signals, since the data for their underrepresented classes are more sparse.

It is important to note that in a real-world setting developing 31 task-specific models is obviously significantly more expensive, both in terms of training and deployment. In many practical situations we would still favour a more computationally economical solution even if the alternative has a marginal quality improvement. Another important comment is that even though MTL ``as is" performance on the tasks is on average lower than the optimally tuned models, they still significantly outperform basic benchmarks (majority class prediction for classifications and mean value for regressions) across every single task.

\section{Conclusion}

This work proposes a scalable deep learning solution to the problem of finding product representations based on the products' image galleries. We use multi-task learning as the main mechanism to enforce the inductive bias in favour of learning generalizable representations. We demonstrate how our method outperforms two strong baselines on a new downstream task. In addition we analyze comparative performance of the two MTL approaches which we test on the training tasks, and show that in case of binary classification tasks there is a strong relationship between label distribution skew and relative performance of the MTL against task-specific models. This finding suggests that MTL can be a useful tool for dealing with sparsity for low-resource tasks. To the best of our knowledge, this work offers the first assessment of MTL approach to building gallery-based product representations in a large-scale industrial setting.

\begin{acks}
The authors wish to thank Shlomi Haver, Stas Girkin and Ivana Rebic for their help, as well as Ron Rostovsky, Ron Reish, Rubens Sonntag, Noa Barbiro and Nitzan Mekel-Bobrov for their support and insightful discussions.
\end{acks}

\bibliographystyle{ACM-Reference-Format}
\bibliography{gallery_rep}


\begin{thebibliography}{34}


\ifx \showCODEN    \undefined \def \showCODEN     #1{\unskip}     \fi
\ifx \showDOI      \undefined \def \showDOI       #1{#1}\fi
\ifx \showISBNx    \undefined \def \showISBNx     #1{\unskip}     \fi
\ifx \showISBNxiii \undefined \def \showISBNxiii  #1{\unskip}     \fi
\ifx \showISSN     \undefined \def \showISSN      #1{\unskip}     \fi
\ifx \showLCCN     \undefined \def \showLCCN      #1{\unskip}     \fi
\ifx \shownote     \undefined \def \shownote      #1{#1}          \fi
\ifx \showarticletitle \undefined \def \showarticletitle #1{#1}   \fi
\ifx \showURL      \undefined \def \showURL       {\relax}        \fi
\providecommand\bibfield[2]{#2}
\providecommand\bibinfo[2]{#2}
\providecommand\natexlab[1]{#1}
\providecommand\showeprint[2][]{arXiv:#2}

\bibitem[\protect\citeauthoryear{Abadi, Agarwal, Barham, Brevdo, Chen, Citro,
  Corrado, Davis, Dean, Devin, Ghemawat, Goodfellow, Harp, Irving, Isard, Jia,
  Jozefowicz, Kaiser, Kudlur, Levenberg, Man\'{e}, Monga, Moore, Murray, Olah,
  Schuster, Shlens, Steiner, Sutskever, Talwar, Tucker, Vanhoucke, Vasudevan,
  Vi\'{e}gas, Vinyals, Warden, Wattenberg, Wicke, Yu, and Zheng}{Abadi
  et~al\mbox{.}}{2015}]%
        {tensorflow2015-whitepaper}
\bibfield{author}{\bibinfo{person}{Mart\'{\i}n Abadi}, \bibinfo{person}{Ashish
  Agarwal}, \bibinfo{person}{Paul Barham}, \bibinfo{person}{Eugene Brevdo},
  \bibinfo{person}{Zhifeng Chen}, \bibinfo{person}{Craig Citro},
  \bibinfo{person}{Greg~S. Corrado}, \bibinfo{person}{Andy Davis},
  \bibinfo{person}{Jeffrey Dean}, \bibinfo{person}{Matthieu Devin},
  \bibinfo{person}{Sanjay Ghemawat}, \bibinfo{person}{Ian Goodfellow},
  \bibinfo{person}{Andrew Harp}, \bibinfo{person}{Geoffrey Irving},
  \bibinfo{person}{Michael Isard}, \bibinfo{person}{Yangqing Jia},
  \bibinfo{person}{Rafal Jozefowicz}, \bibinfo{person}{Lukasz Kaiser},
  \bibinfo{person}{Manjunath Kudlur}, \bibinfo{person}{Josh Levenberg},
  \bibinfo{person}{Dan Man\'{e}}, \bibinfo{person}{Rajat Monga},
  \bibinfo{person}{Sherry Moore}, \bibinfo{person}{Derek Murray},
  \bibinfo{person}{Chris Olah}, \bibinfo{person}{Mike Schuster},
  \bibinfo{person}{Jonathon Shlens}, \bibinfo{person}{Benoit Steiner},
  \bibinfo{person}{Ilya Sutskever}, \bibinfo{person}{Kunal Talwar},
  \bibinfo{person}{Paul Tucker}, \bibinfo{person}{Vincent Vanhoucke},
  \bibinfo{person}{Vijay Vasudevan}, \bibinfo{person}{Fernanda Vi\'{e}gas},
  \bibinfo{person}{Oriol Vinyals}, \bibinfo{person}{Pete Warden},
  \bibinfo{person}{Martin Wattenberg}, \bibinfo{person}{Martin Wicke},
  \bibinfo{person}{Yuan Yu}, {and} \bibinfo{person}{Xiaoqiang Zheng}.}
  \bibinfo{year}{2015}\natexlab{}.
\newblock \bibinfo{title}{{TensorFlow}: Large-Scale Machine Learning on
  Heterogeneous Systems}.
\newblock
\newblock
\urldef\tempurl%
\url{http://tensorflow.org/}
\showURL{%
\tempurl}
\newblock
\shownote{Software available from tensorflow.org.}


\bibitem[\protect\citeauthoryear{Caruana}{Caruana}{1997}]%
        {caruana1997multitask}
\bibfield{author}{\bibinfo{person}{Rich Caruana}.}
  \bibinfo{year}{1997}\natexlab{}.
\newblock \showarticletitle{Multitask learning}.
\newblock \bibinfo{journal}{\emph{Machine learning}} \bibinfo{volume}{28},
  \bibinfo{number}{1} (\bibinfo{year}{1997}), \bibinfo{pages}{41--75}.
\newblock


\bibitem[\protect\citeauthoryear{Deng, Dong, Socher, Li, Li, and Fei-Fei}{Deng
  et~al\mbox{.}}{2009}]%
        {imagenet_cvpr09}
\bibfield{author}{\bibinfo{person}{J. Deng}, \bibinfo{person}{W. Dong},
  \bibinfo{person}{R. Socher}, \bibinfo{person}{L.-J. Li}, \bibinfo{person}{K.
  Li}, {and} \bibinfo{person}{L. Fei-Fei}.} \bibinfo{year}{2009}\natexlab{}.
\newblock \showarticletitle{{ImageNet: A Large-Scale Hierarchical Image
  Database}}. In \bibinfo{booktitle}{\emph{CVPR09}}.
\newblock


\bibitem[\protect\citeauthoryear{Devlin, Chang, Lee, and Toutanova}{Devlin
  et~al\mbox{.}}{2018}]%
        {devlin2018bert}
\bibfield{author}{\bibinfo{person}{Jacob Devlin}, \bibinfo{person}{Ming-Wei
  Chang}, \bibinfo{person}{Kenton Lee}, {and} \bibinfo{person}{Kristina
  Toutanova}.} \bibinfo{year}{2018}\natexlab{}.
\newblock \showarticletitle{Bert: Pre-training of deep bidirectional
  transformers for language understanding}.
\newblock \bibinfo{journal}{\emph{arXiv preprint arXiv:1810.04805}}
  (\bibinfo{year}{2018}).
\newblock


\bibitem[\protect\citeauthoryear{Di, Sundaresan, Piramuthu, and Bhardwaj}{Di
  et~al\mbox{.}}{2014}]%
        {di2014picture}
\bibfield{author}{\bibinfo{person}{Wei Di}, \bibinfo{person}{Neel Sundaresan},
  \bibinfo{person}{Robinson Piramuthu}, {and} \bibinfo{person}{Anurag
  Bhardwaj}.} \bibinfo{year}{2014}\natexlab{}.
\newblock \showarticletitle{Is a picture really worth a thousand words? -on the
  role of images in e-commerce}. In \bibinfo{booktitle}{\emph{Proceedings of
  the 7th ACM international conference on Web search and data mining}}.
  \bibinfo{pages}{633--642}.
\newblock


\bibitem[\protect\citeauthoryear{Duong, Cohn, Bird, and Cook}{Duong
  et~al\mbox{.}}{2015}]%
        {duong-etal-2015-low}
\bibfield{author}{\bibinfo{person}{Long Duong}, \bibinfo{person}{Trevor Cohn},
  \bibinfo{person}{Steven Bird}, {and} \bibinfo{person}{Paul Cook}.}
  \bibinfo{year}{2015}\natexlab{}.
\newblock \showarticletitle{Low Resource Dependency Parsing: Cross-lingual
  Parameter Sharing in a Neural Network Parser}. In
  \bibinfo{booktitle}{\emph{Proceedings of the 53rd Annual Meeting of the
  Association for Computational Linguistics and the 7th International Joint
  Conference on Natural Language Processing (Volume 2: Short Papers)}}.
  \bibinfo{publisher}{Association for Computational Linguistics},
  \bibinfo{address}{Beijing, China}, \bibinfo{pages}{845--850}.
\newblock
\urldef\tempurl%
\url{https://doi.org/10.3115/v1/P15-2139}
\showDOI{\tempurl}


\bibitem[\protect\citeauthoryear{Eigen and Fergus}{Eigen and Fergus}{2015}]%
        {eigen2015predicting}
\bibfield{author}{\bibinfo{person}{David Eigen} {and} \bibinfo{person}{Rob
  Fergus}.} \bibinfo{year}{2015}\natexlab{}.
\newblock \showarticletitle{Predicting depth, surface normals and semantic
  labels with a common multi-scale convolutional architecture}. In
  \bibinfo{booktitle}{\emph{Proceedings of the IEEE international conference on
  computer vision}}. \bibinfo{pages}{2650--2658}.
\newblock


\bibitem[\protect\citeauthoryear{Gong, Lee, Stephenson, Renduchintala, Padhy,
  Ndirango, Keskin, and Elibol}{Gong et~al\mbox{.}}{2019}]%
        {gong2019comparison}
\bibfield{author}{\bibinfo{person}{Ting Gong}, \bibinfo{person}{Tyler Lee},
  \bibinfo{person}{Cory Stephenson}, \bibinfo{person}{Venkata Renduchintala},
  \bibinfo{person}{Suchismita Padhy}, \bibinfo{person}{Anthony Ndirango},
  \bibinfo{person}{Gokce Keskin}, {and} \bibinfo{person}{Oguz~H Elibol}.}
  \bibinfo{year}{2019}\natexlab{}.
\newblock \showarticletitle{A Comparison of Loss Weighting Strategies for Multi
  task Learning in Deep Neural Networks}.
\newblock \bibinfo{journal}{\emph{IEEE Access}}  \bibinfo{volume}{7}
  (\bibinfo{year}{2019}), \bibinfo{pages}{141627--141632}.
\newblock


\bibitem[\protect\citeauthoryear{He, Zhang, Ren, and Sun}{He
  et~al\mbox{.}}{2016}]%
        {he2016deep}
\bibfield{author}{\bibinfo{person}{Kaiming He}, \bibinfo{person}{Xiangyu
  Zhang}, \bibinfo{person}{Shaoqing Ren}, {and} \bibinfo{person}{Jian Sun}.}
  \bibinfo{year}{2016}\natexlab{}.
\newblock \showarticletitle{Deep residual learning for image recognition}. In
  \bibinfo{booktitle}{\emph{Proceedings of the IEEE conference on computer
  vision and pattern recognition}}. \bibinfo{pages}{770--778}.
\newblock


\bibitem[\protect\citeauthoryear{Indyk and Motwani}{Indyk and Motwani}{1998}]%
        {indyk1998approximate}
\bibfield{author}{\bibinfo{person}{Piotr Indyk} {and} \bibinfo{person}{Rajeev
  Motwani}.} \bibinfo{year}{1998}\natexlab{}.
\newblock \showarticletitle{Approximate nearest neighbors: towards removing the
  curse of dimensionality}. In \bibinfo{booktitle}{\emph{Proceedings of the
  thirtieth annual ACM symposium on Theory of computing}}.
  \bibinfo{pages}{604--613}.
\newblock


\bibitem[\protect\citeauthoryear{Kalra, Adnan, Taylor, and Tizhoosh}{Kalra
  et~al\mbox{.}}{2019}]%
        {kalra2019learning}
\bibfield{author}{\bibinfo{person}{Shivam Kalra}, \bibinfo{person}{Mohammed
  Adnan}, \bibinfo{person}{Graham Taylor}, {and} \bibinfo{person}{Hamid
  Tizhoosh}.} \bibinfo{year}{2019}\natexlab{}.
\newblock \showarticletitle{Learning Permutation Invariant Representations
  using Memory Networks}.
\newblock \bibinfo{journal}{\emph{arXiv preprint arXiv:1911.07984}}
  (\bibinfo{year}{2019}).
\newblock


\bibitem[\protect\citeauthoryear{Karpathy}{Karpathy}{2019}]%
        {karpathy}
\bibfield{author}{\bibinfo{person}{Andej Karpathy}.}
  \bibinfo{year}{2019}\natexlab{}.
\newblock \bibinfo{booktitle}{\emph{Multi-task learning in the wilderness}}.
\newblock
\urldef\tempurl%
\url{https://slideslive.com/38917690/multitask-learning-in-the-wilderness}
\showURL{%
\tempurl}


\bibitem[\protect\citeauthoryear{Kendall, Gal, and Cipolla}{Kendall
  et~al\mbox{.}}{2018}]%
        {kendall2018multi}
\bibfield{author}{\bibinfo{person}{Alex Kendall}, \bibinfo{person}{Yarin Gal},
  {and} \bibinfo{person}{Roberto Cipolla}.} \bibinfo{year}{2018}\natexlab{}.
\newblock \showarticletitle{Multi-task learning using uncertainty to weigh
  losses for scene geometry and semantics}. In
  \bibinfo{booktitle}{\emph{Proceedings of the IEEE Conference on Computer
  Vision and Pattern Recognition}}. \bibinfo{pages}{7482--7491}.
\newblock


\bibitem[\protect\citeauthoryear{{Kingma} and {Ba}}{{Kingma} and {Ba}}{2014}]%
        {kingma2014method}
\bibfield{author}{\bibinfo{person}{Diederik~P. {Kingma}} {and}
  \bibinfo{person}{Jimmy {Ba}}.} \bibinfo{year}{2014}\natexlab{}.
\newblock \showarticletitle{Adam: A Method for Stochastic Optimization}. In
  \bibinfo{booktitle}{\emph{Proceedings of the 3rd International Conference on
  Learning Representations (ICLR)}}.
\newblock
\urldef\tempurl%
\url{http://arxiv.org/abs/1412.6980}
\showURL{%
\tempurl}


\bibitem[\protect\citeauthoryear{Kokkinos}{Kokkinos}{2017}]%
        {kokkinos2017ubernet}
\bibfield{author}{\bibinfo{person}{Iasonas Kokkinos}.}
  \bibinfo{year}{2017}\natexlab{}.
\newblock \showarticletitle{Ubernet: Training a universal convolutional neural
  network for low-, mid-, and high-level vision using diverse datasets and
  limited memory}. In \bibinfo{booktitle}{\emph{Proceedings of the IEEE
  Conference on Computer Vision and Pattern Recognition}}.
  \bibinfo{pages}{6129--6138}.
\newblock


\bibitem[\protect\citeauthoryear{Korshunova, Degrave, Husz{\'a}r, Gal, Gretton,
  and Dambre}{Korshunova et~al\mbox{.}}{2018}]%
        {korshunova2018bruno}
\bibfield{author}{\bibinfo{person}{Iryna Korshunova}, \bibinfo{person}{Jonas
  Degrave}, \bibinfo{person}{Ferenc Husz{\'a}r}, \bibinfo{person}{Yarin Gal},
  \bibinfo{person}{Arthur Gretton}, {and} \bibinfo{person}{Joni Dambre}.}
  \bibinfo{year}{2018}\natexlab{}.
\newblock \showarticletitle{Bruno: A deep recurrent model for exchangeable
  data}. In \bibinfo{booktitle}{\emph{Advances in Neural Information Processing
  Systems}}. \bibinfo{pages}{7190--7198}.
\newblock


\bibitem[\protect\citeauthoryear{Krizhevsky, Sutskever, and Hinton}{Krizhevsky
  et~al\mbox{.}}{2012}]%
        {krizhevsky2012imagenet}
\bibfield{author}{\bibinfo{person}{Alex Krizhevsky}, \bibinfo{person}{Ilya
  Sutskever}, {and} \bibinfo{person}{Geoffrey~E Hinton}.}
  \bibinfo{year}{2012}\natexlab{}.
\newblock \showarticletitle{Imagenet classification with deep convolutional
  neural networks}. In \bibinfo{booktitle}{\emph{Advances in neural information
  processing systems}}. \bibinfo{pages}{1097--1105}.
\newblock


\bibitem[\protect\citeauthoryear{Lee, Lee, Kim, Kosiorek, Choi, and Teh}{Lee
  et~al\mbox{.}}{2019}]%
        {lee2019set}
\bibfield{author}{\bibinfo{person}{Juho Lee}, \bibinfo{person}{Yoonho Lee},
  \bibinfo{person}{Jungtaek Kim}, \bibinfo{person}{Adam Kosiorek},
  \bibinfo{person}{Seungjin Choi}, {and} \bibinfo{person}{Yee~Whye Teh}.}
  \bibinfo{year}{2019}\natexlab{}.
\newblock \showarticletitle{Set Transformer: A Framework for Attention-based
  Permutation-Invariant Neural Networks}. In
  \bibinfo{booktitle}{\emph{International Conference on Machine Learning}}.
  \bibinfo{pages}{3744--3753}.
\newblock


\bibitem[\protect\citeauthoryear{Ma, Mezghani, Wilber, Hong, Piramuthu, Naaman,
  and Belongie}{Ma et~al\mbox{.}}{2019}]%
        {ma2019understanding}
\bibfield{author}{\bibinfo{person}{Xiao Ma}, \bibinfo{person}{Lina Mezghani},
  \bibinfo{person}{Kimberly Wilber}, \bibinfo{person}{Hui Hong},
  \bibinfo{person}{Robinson Piramuthu}, \bibinfo{person}{Mor Naaman}, {and}
  \bibinfo{person}{Serge Belongie}.} \bibinfo{year}{2019}\natexlab{}.
\newblock \showarticletitle{Understanding Image Quality and Trust in
  Peer-to-Peer Marketplaces}. In \bibinfo{booktitle}{\emph{2019 IEEE Winter
  Conference on Applications of Computer Vision (WACV)}}. IEEE,
  \bibinfo{pages}{511--520}.
\newblock


\bibitem[\protect\citeauthoryear{McInnes, Healy, and Melville}{McInnes
  et~al\mbox{.}}{2018}]%
        {mcinnes2018umap}
\bibfield{author}{\bibinfo{person}{Leland McInnes}, \bibinfo{person}{John
  Healy}, {and} \bibinfo{person}{James Melville}.}
  \bibinfo{year}{2018}\natexlab{}.
\newblock \showarticletitle{Umap: Uniform manifold approximation and projection
  for dimension reduction}.
\newblock \bibinfo{journal}{\emph{arXiv preprint arXiv:1802.03426}}
  (\bibinfo{year}{2018}).
\newblock


\bibitem[\protect\citeauthoryear{Misra, Shrivastava, Gupta, and Hebert}{Misra
  et~al\mbox{.}}{2016}]%
        {misra2016cross}
\bibfield{author}{\bibinfo{person}{Ishan Misra}, \bibinfo{person}{Abhinav
  Shrivastava}, \bibinfo{person}{Abhinav Gupta}, {and} \bibinfo{person}{Martial
  Hebert}.} \bibinfo{year}{2016}\natexlab{}.
\newblock \showarticletitle{Cross-stitch networks for multi-task learning}. In
  \bibinfo{booktitle}{\emph{Proceedings of the IEEE Conference on Computer
  Vision and Pattern Recognition}}. \bibinfo{pages}{3994--4003}.
\newblock


\bibitem[\protect\citeauthoryear{Oquab, Bottou, Laptev, and Sivic}{Oquab
  et~al\mbox{.}}{2014}]%
        {oquab2014learning}
\bibfield{author}{\bibinfo{person}{Maxime Oquab}, \bibinfo{person}{Leon
  Bottou}, \bibinfo{person}{Ivan Laptev}, {and} \bibinfo{person}{Josef Sivic}.}
  \bibinfo{year}{2014}\natexlab{}.
\newblock \showarticletitle{Learning and transferring mid-level image
  representations using convolutional neural networks}. In
  \bibinfo{booktitle}{\emph{Proceedings of the IEEE conference on computer
  vision and pattern recognition}}. \bibinfo{pages}{1717--1724}.
\newblock


\bibitem[\protect\citeauthoryear{Ren, He, Girshick, and Sun}{Ren
  et~al\mbox{.}}{2015}]%
        {ren2015faster}
\bibfield{author}{\bibinfo{person}{Shaoqing Ren}, \bibinfo{person}{Kaiming He},
  \bibinfo{person}{Ross Girshick}, {and} \bibinfo{person}{Jian Sun}.}
  \bibinfo{year}{2015}\natexlab{}.
\newblock \showarticletitle{Faster r-cnn: Towards real-time object detection
  with region proposal networks}. In \bibinfo{booktitle}{\emph{Advances in
  neural information processing systems}}. \bibinfo{pages}{91--99}.
\newblock


\bibitem[\protect\citeauthoryear{Ruder}{Ruder}{2017}]%
        {ruder2017overview}
\bibfield{author}{\bibinfo{person}{Sebastian Ruder}.}
  \bibinfo{year}{2017}\natexlab{}.
\newblock \showarticletitle{An overview of multi-task learning in deep neural
  networks}.
\newblock \bibinfo{journal}{\emph{arXiv preprint arXiv:1706.05098}}
  (\bibinfo{year}{2017}).
\newblock


\bibitem[\protect\citeauthoryear{Sener and Koltun}{Sener and Koltun}{2018}]%
        {sener2018multi}
\bibfield{author}{\bibinfo{person}{Ozan Sener} {and} \bibinfo{person}{Vladlen
  Koltun}.} \bibinfo{year}{2018}\natexlab{}.
\newblock \showarticletitle{Multi-task learning as multi-objective
  optimization}. In \bibinfo{booktitle}{\emph{Advances in Neural Information
  Processing Systems}}. \bibinfo{pages}{527--538}.
\newblock


\bibitem[\protect\citeauthoryear{Simonyan and Zisserman}{Simonyan and
  Zisserman}{2014}]%
        {simonyan2014very}
\bibfield{author}{\bibinfo{person}{Karen Simonyan} {and}
  \bibinfo{person}{Andrew Zisserman}.} \bibinfo{year}{2014}\natexlab{}.
\newblock \showarticletitle{Very deep convolutional networks for large-scale
  image recognition}.
\newblock \bibinfo{journal}{\emph{arXiv preprint arXiv:1409.1556}}
  (\bibinfo{year}{2014}).
\newblock


\bibitem[\protect\citeauthoryear{Standley, Zamir, Chen, Guibas, Malik, and
  Savarese}{Standley et~al\mbox{.}}{2019}]%
        {standley2019tasks}
\bibfield{author}{\bibinfo{person}{Trevor Standley}, \bibinfo{person}{Amir~R
  Zamir}, \bibinfo{person}{Dawn Chen}, \bibinfo{person}{Leonidas Guibas},
  \bibinfo{person}{Jitendra Malik}, {and} \bibinfo{person}{Silvio Savarese}.}
  \bibinfo{year}{2019}\natexlab{}.
\newblock \showarticletitle{Which Tasks Should Be Learned Together in
  Multi-task Learning?}
\newblock \bibinfo{journal}{\emph{arXiv preprint arXiv:1905.07553}}
  (\bibinfo{year}{2019}).
\newblock


\bibitem[\protect\citeauthoryear{Szegedy, Liu, Jia, Sermanet, Reed, Anguelov,
  Erhan, Vanhoucke, and Rabinovich}{Szegedy et~al\mbox{.}}{2015}]%
        {szegedy2015going}
\bibfield{author}{\bibinfo{person}{Christian Szegedy}, \bibinfo{person}{Wei
  Liu}, \bibinfo{person}{Yangqing Jia}, \bibinfo{person}{Pierre Sermanet},
  \bibinfo{person}{Scott Reed}, \bibinfo{person}{Dragomir Anguelov},
  \bibinfo{person}{Dumitru Erhan}, \bibinfo{person}{Vincent Vanhoucke}, {and}
  \bibinfo{person}{Andrew Rabinovich}.} \bibinfo{year}{2015}\natexlab{}.
\newblock \showarticletitle{Going deeper with convolutions}. In
  \bibinfo{booktitle}{\emph{Proceedings of the IEEE conference on computer
  vision and pattern recognition}}. \bibinfo{pages}{1--9}.
\newblock


\bibitem[\protect\citeauthoryear{Teichmann, Weber, Zoellner, Cipolla, and
  Urtasun}{Teichmann et~al\mbox{.}}{2018}]%
        {teichmann2018multinet}
\bibfield{author}{\bibinfo{person}{Marvin Teichmann}, \bibinfo{person}{Michael
  Weber}, \bibinfo{person}{Marius Zoellner}, \bibinfo{person}{Roberto Cipolla},
  {and} \bibinfo{person}{Raquel Urtasun}.} \bibinfo{year}{2018}\natexlab{}.
\newblock \showarticletitle{Multinet: Real-time joint semantic reasoning for
  autonomous driving}. In \bibinfo{booktitle}{\emph{2018 IEEE Intelligent
  Vehicles Symposium (IV)}}. IEEE, \bibinfo{pages}{1013--1020}.
\newblock


\bibitem[\protect\citeauthoryear{Vaswani, Shazeer, Parmar, Uszkoreit, Jones,
  Gomez, Kaiser, and Polosukhin}{Vaswani et~al\mbox{.}}{2017}]%
        {Vaswani2017AttentionIA}
\bibfield{author}{\bibinfo{person}{Ashish Vaswani}, \bibinfo{person}{Noam
  Shazeer}, \bibinfo{person}{Niki Parmar}, \bibinfo{person}{Jakob Uszkoreit},
  \bibinfo{person}{Llion Jones}, \bibinfo{person}{Aidan~N. Gomez},
  \bibinfo{person}{Lukasz Kaiser}, {and} \bibinfo{person}{Illia Polosukhin}.}
  \bibinfo{year}{2017}\natexlab{}.
\newblock \showarticletitle{Attention is All you Need}. In
  \bibinfo{booktitle}{\emph{NIPS}}.
\newblock


\bibitem[\protect\citeauthoryear{Vinyals, Bengio, and Kudlur}{Vinyals
  et~al\mbox{.}}{2015}]%
        {vinyals2015order}
\bibfield{author}{\bibinfo{person}{Oriol Vinyals}, \bibinfo{person}{Samy
  Bengio}, {and} \bibinfo{person}{Manjunath Kudlur}.}
  \bibinfo{year}{2015}\natexlab{}.
\newblock \showarticletitle{Order matters: Sequence to sequence for sets}.
\newblock \bibinfo{journal}{\emph{arXiv preprint arXiv:1511.06391}}
  (\bibinfo{year}{2015}).
\newblock


\bibitem[\protect\citeauthoryear{Weston, Chopra, and Bordes}{Weston
  et~al\mbox{.}}{2014}]%
        {weston2014memory}
\bibfield{author}{\bibinfo{person}{Jason Weston}, \bibinfo{person}{Sumit
  Chopra}, {and} \bibinfo{person}{Antoine Bordes}.}
  \bibinfo{year}{2014}\natexlab{}.
\newblock \showarticletitle{Memory networks}.
\newblock \bibinfo{journal}{\emph{arXiv preprint arXiv:1410.3916}}
  (\bibinfo{year}{2014}).
\newblock


\bibitem[\protect\citeauthoryear{Zaheer, Kottur, Ravanbakhsh, Poczos,
  Salakhutdinov, and Smola}{Zaheer et~al\mbox{.}}{2017}]%
        {zaheer2017deep}
\bibfield{author}{\bibinfo{person}{Manzil Zaheer}, \bibinfo{person}{Satwik
  Kottur}, \bibinfo{person}{Siamak Ravanbakhsh}, \bibinfo{person}{Barnabas
  Poczos}, \bibinfo{person}{Ruslan~R Salakhutdinov}, {and}
  \bibinfo{person}{Alexander~J Smola}.} \bibinfo{year}{2017}\natexlab{}.
\newblock \showarticletitle{Deep sets}. In \bibinfo{booktitle}{\emph{Advances
  in neural information processing systems}}. \bibinfo{pages}{3391--3401}.
\newblock


\bibitem[\protect\citeauthoryear{Zhang and Yang}{Zhang and Yang}{2018}]%
        {zhang2018overview}
\bibfield{author}{\bibinfo{person}{Yu Zhang} {and} \bibinfo{person}{Qiang
  Yang}.} \bibinfo{year}{2018}\natexlab{}.
\newblock \showarticletitle{An overview of multi-task learning}.
\newblock \bibinfo{journal}{\emph{National Science Review}}
  \bibinfo{volume}{5}, \bibinfo{number}{1} (\bibinfo{year}{2018}),
  \bibinfo{pages}{30--43}.
\newblock


\end{thebibliography}

\end{document}